\documentclass{article}

\usepackage[preprint]{unireps_2025}

\usepackage[utf8]{inputenc} %
\usepackage[T1]{fontenc}    %
\usepackage{hyperref}       %
\usepackage{url}            %
\usepackage{booktabs}
\usepackage{tabularx}%
\usepackage{amsfonts}       %
\usepackage{nicefrac}       %
\usepackage{microtype}      %
\usepackage[dvipsnames]{xcolor} %
\usepackage{natbib}
\usepackage{array,booktabs}
\usepackage{amsmath}
\usepackage{bbding}
\usepackage{txfonts}
\usepackage{pifont}
\usepackage{graphicx}
\usepackage{textcomp}
\usepackage{float}
\usepackage{longtable, booktabs}
\title{Towards Interpretable Deep Neural Networks for Tabular Data}

\author{%
  Khawla Elhadri\\
  Marburg University\\
  \texttt{khawla.elhadri@uni-marburg.de} \\
   \And
   Jörg Schlötterer \\
   Marburg University\\
   \texttt{joerg.schloetterer@uni-marburg.de} \\
   \AND
   Christin Seifert \\
   Marburg University\\
   \texttt{christin.seifert@uni-marburg.de} \\
}

\newcommand {\otoprule}{\midrule [\heavyrulewidth]}
\newcolumntype {+}{ >{\global\let\currentrowstyle\relax}}
\newcolumntype {^}{ >{\currentrowstyle }}
 \newcommand {\rowstyle}[1]{\gdef\currentrowstyle{#1} %
 #1\ignorespaces
 }
\newcommand{\tabhead}{\rowstyle{\bfseries}}

\usepackage{xspace}
\newcommand{\modelname}{\textsc{XNNTab}\xspace}

\newcommand{\dsadult}{\textsc{ADULT}\xspace}
\newcommand{\dschurn}{\textsc{CHURN}\xspace}

\usepackage{xcolor} %

\begin{document}

\maketitle

\begin{abstract}
Tabular data is the foundation of many applications in fields such as finance and healthcare. Although DNNs tailored for tabular data achieve competitive predictive performance, they are blackboxes with little interpretability. We introduce \modelname, a neural architecture that uses a sparse autoencoder (SAE) to learn a dictionary of monosemantic features within the latent space used for prediction.
Using an automated method, we assign human-interpretable semantics to these features. This allows us to represent predictions as linear combinations of semantically meaningful components.
Empirical evaluations demonstrate that \modelname attains performance comparable to that of state-of-the-art, black-box neural models and classical machine learning approaches while being fully interpretable. 

\end{abstract}

\section{Introduction}
\label{sec:intro}
Tabular data is the most common type of data in a wide range of industries, including advertising, finance and healthcare.  While deep  neural networks (DNNs) achieved major advances in computer vision, their performance on tabular data remained below that of Gradient Boosted Decision Trees (GBDTs)~\citep{Chen_2016_XGBoostScalableTree, Khan_2022_TransformersVisionSurvey}. To improve the performance of DNNs on tabular data, several deep learning methods have been developed with specialized architectures that take into account the unique traits of tabular data (mixture of categorical, ordinal and continuous features) in order to be on par~\citep{Huang_2020_TabTransformerTabularData} or even outperform GBDTs~\citep{Chen_2024_CanDeepLearning}.
While these methods address the limitations of DNNs in terms of performance, DNNs still remain blackboxes, whose decisions are hard to communicate to end users~\citep{Schwalbe2023_ComprehensiveTaxonomyExplainable}.

In this paper, we address the interpretability limitation of existing DNNs and present \modelname, a deep neural network with a sparse autoencoder (SAE) component. The SAE learns a dictionary of monosemantic, sparse features that are used for outcome prediction. We take inspiration from the success of SAEs in learning interpretable features from LLM's internal activations~\citep{Huben_2023_SparseAutoencodersFind, yun_2021_transformervisualization}.
We use an automated method to assign human-understandable semantics to the learned monosemantic dictionary features. Our method is illustrated in Figure~\ref{fig:method-overview} and described in Section~\ref{sec:method}.
We show that \modelname outperforms interpretable  models and has similar performance to classical blackbox models. Additionally,  \modelname is fully interpretable: the final prediction is a simple linear combination of the dictionary features, which have easily accessible human-understandable semantics.

\section{Related Work}
\label{sec:relatedwork}

Gradient boosted decision trees (GBDTs)~\citep{Chen_2016_XGBoostScalableTree,Khan_2022_TransformersVisionSurvey} have shown state-of-the art performance on tabular data~\citep{Grinsztajn_2022_Whytreebasedmodels}. To close this performance gap to GBDTs, neural architectures specifically designed for tabular data have been proposed, such as attention-based architectures~\citep{Gorishniy_2021_RevisitingDeepLearning, Somepalli_2022_SAINTImprovedNeural, Yan_2023_T2GFORMERorganizingtabular} or retrieval-augmented models~\citep{Somepalli_2022_SAINTImprovedNeural, Gorishniy_2023_TabRTabularDeepa}. Additional techniques include tailored regularizations~\citep{Jeffares_2023_TANGOSRegularizingTabular}, deep ensembles~\citep{Gorishniy_2024_TabMAdvancingtabular}, attentive feature selection strategies~\citep{Arik_2021_TabNetAttentiveInterpretable}, categorical feature embedding~\citep{, Huang_2020_TabTransformerTabularData} and reconstruction of binned feature indices~\citep{Lee_2024_Binningpretexttask}.

Recent methods focus not only on performance, but also aim to include interpretability.
ProtoGate~\citep{Jiang_2024_ProtoGateprototypebasedneural} is an ante-hoc interpretable architecture for tabular data, which learns prototypes of classes and predicts new instances based on their similarity to class prototypes. InterpreTabNet~\citep{Si_2024_InterpreTabNetdistillingpredictive}, a variant of TabNet~\citep{Arik_2021_TabNetAttentiveInterpretable} adds post-hoc explanations by learning sparse feature attribution masks and using LLMs to interpret the learned features from the masks.
Similar to ProtoGate and different from InterpetTabNet,  \modelname is intrinsically interpretable, i.e., it uses interpretable features directly for prediction. While ProtoGate relies on class prototypes, \modelname learns relevant feature combinations that describe parts of instances, similar to part-prototype models~\citep{elhadri_2025_lookslikewhatchallenges}.

\section{Method}
\label{sec:method}

\begin{figure*}
    \centering
    \includegraphics[width=\linewidth]{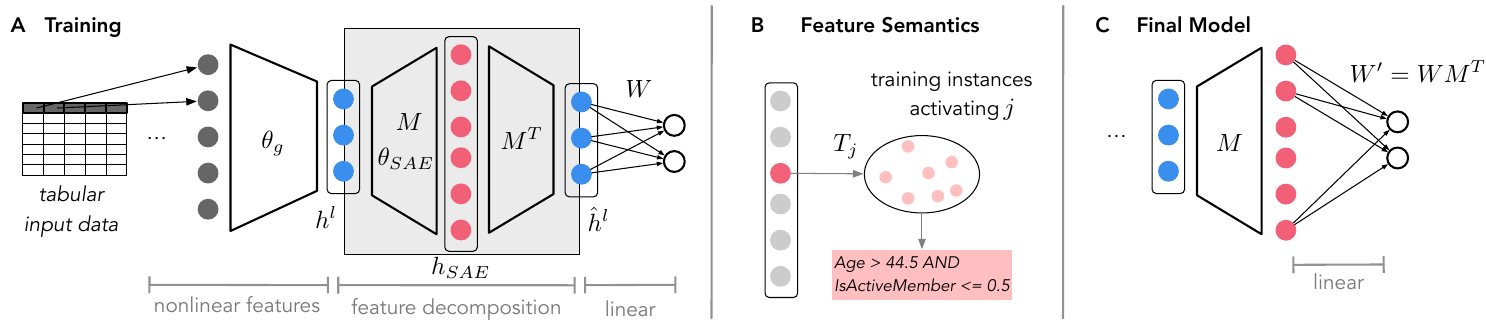}
    \caption{Architecture overview. \textbf{A.} An MLP learns nonlinear features, which are decomposed into monosemantic dictionary features using an SAE. \textbf{B.} Dictionary features are assigned a human-interpretable meaning by learning rules for the subset of training instances that highly activate a specific feature. \textbf{C.} Predictions are linear combinations of monosemantic features by combining the linear model components (the decoder, $M^T$, and the linear layer of the MLP, $W$).}
    \label{fig:method-overview}
\end{figure*}

\modelname consists of a standard neural blackbox (e.g., an MLP) to learn nonlinear latent features, and a sparse autoencoder (SAE)~\citep{Huben_2023_SparseAutoencodersFind} to decompose this latent polysemantic feature space into monosemantic features (cf. Figure~\ref{fig:method-overview}, A). The monosemantic features are in a latent space with unknown (but unique) semantics. To assign meaning to each feature, we learn rules that describe the subset of the training samples that highly activate each feature (cf. Figure~\ref{fig:method-overview}, B). Finally, we combine the last linear layers into a single linear layer by multiplying their weight matrices (cf. Figure~\ref{fig:method-overview}, C). 

\subsection{Architecture}
The base blackbox model is an MLP with multiple hidden layers and a softmax output. We denote the penultimate layer as $l$, and the representations of this layer as $h_l$.  
The MLP learns a function $\hat{y} = f(x)$. $f(x)$ can be decomposed into a function $g$ learning the hidden representations  $h_l$ and a linear model $h$ predicting the target based on $h_l$, i.e., $\hat{y} = h(h_l)=h(g(x))$. $g(\cdot)$ is characterized by the parameters $\theta_g$, and $h(\cdot)$ is characterized by the weight matrix $W$.

We adopt the SAE architecture introduced by~\citet{Huben_2023_SparseAutoencodersFind}.
The SAE consists of an input layer, a hidden layer, and an output layer. The weights of the encoder and decoder are shared, and given by the matrices $M$ and $M^T$, respectively. The encoder has an additional bias vector $b$, the decoder has not~\citep{Huben_2023_SparseAutoencodersFind}. We set the size of the hidden layer as $d_{hid} = R\cdot d_{in}$ with $d_{in}$ being the dimension of the MLP's layer $l$. $R$ is a hyperparameter that controls the size of the feature dictionary.
The input to the SAE are the hidden representations of the blackbox MLP $h_l$.  The output of the SAE is given by 
$\hat{h}_l = M^T \Bigl(ReLU(M h_l + b)\Bigr)$.
By design, $\hat{h}_l$ is a sparse linear combination of (latent) dictionary features~\citep{Huben_2023_SparseAutoencodersFind}.

\subsection{Training}
\modelname is trained in four steps: i) training of the MLP's representation, ii) training of the SAE, and iii) finetuning of the decision layer, iv) combining linear components. 

\textbf{Step 1: MLP training.} The MLP's parameters $\theta_\text{MLP}$= ($\theta_g$, $W$) are trained using cross-entropy loss and L1 regularization. The SAE is not present in this step.

\textbf{Step 2: SAE Training.} To learn a dictionary of monosemantic features we follow the training procedure by \citet{Huben_2023_SparseAutoencodersFind}. The SAE is trained to reconstruct the hidden activations $h_l$ of training samples $x$. The SAE uses a reconstructing loss and a sparsity loss on its hidden activations $h_{SAE}$ to learn sparse dictionary features.

\textbf{Step 3: Finetuning the decision layer.} 
We freeze the parameters $\theta_g$ of the MLP and $\theta_{SAE}$ of the SAE and finetune the weights $W$ in the decision layer with the same loss as in step 1 to make predictions from the reconstructions learned by the SAE $\hat{h}_l$.

\textbf{Step 4: Aggregating linear components.} Both, the decoder part of the SAE and the final decision layer of the MLP are linear layers, and can be combined into one layer. 
Therefore, we directly connect the hidden layer of the SAE to the output and set the weights of this layer to $W' = WM^T$.\footnote{$\hat{h}_l = M^T h_{SAE}$, and $\hat{y} = W \hat{h}_l = W M^T h_{SAE}$ }

\subsection{Learning Representation Semantics}
The hidden representations of the SAE represent monosemantic dictionary features in a latent space, but their semantics are unknown. To assign human-understandable semantics to those features, we follow a similar idea as in~\citep{Huben_2023_SparseAutoencodersFind} for tabular data instead of text (cf. Figure~\ref{fig:method-overview}, B). 
For each dimension in the latent representation of the SAE $j \in \{1, ...d_{hid} \}$, we identify the subset of training samples $T_j$ which highly active this feature, i.e., whose activations are above a threshold $t$. We then use a rule-based classifier to learn simple decision rules describing this subset.

\section{Experiments}
\label{sec:experiments}

\textbf{Datasets and Evaluation Metrics.}
We evaluate on two benchmark datasets for structured data, Adult (\dsadult)~\citep{Vanschoren_2013_OpenML} and Churn Modelling (\dschurn)\footnote{https://www.kaggle.com/datasets/shrutimechlearn/churn-modelling}.
\dsadult contains 15 features describing approx. 50k instances. 
\dschurn contains 14 features and 10k instances. In both cases, the task is binary classification.
For evaluation, we use 5-fold-cross-validation. The training/validation/test proportion of the datasets for each split are 65\%/15\%/20\%. We report accuracy and macro F1 on the test set, averaged across all folds.

\textbf{Models and Parameters.}
We manually define the neural architecture for the MLP and SAE components of \modelname. 
The base MLP has 3 layers with (100, 64, 32) neurons, the SAE $2\times 32= 64$ (R=2) neurons in the hidden layer. 
See Appendix~\ref{appendix:impl_details:paramter_settings} for details on all hyperparameters.

\textbf{Representation Semantics.}
To assign human-understandable semantics to the dictionary features, we employ Skope-rules\footnote{https://github.com/scikit-learn-contrib/skope-rules. Accessed July 2025}.  
For each dictionary feature $f_j$, we retrieve a set $T_j$ of instances for which the activation of neuron $j$ is above the threshold $t=0.75$. We assign $label = 1$ to instances in $T_j$ and $label = 0$ to the remaining instances. The Skope-rules classifier learns a set of decision rules describing instances of the positive class. We set the precision hyperparameter of Skope-rules to 1 and keep the rest of the hyperparameter to the default settings. For each $T_j$, we keep the rule with the highest coverage (recall).%

\begin{figure}[t]
    \centering
    \begin{minipage}[t]{0.5\textwidth}
    \vspace{0pt} %
        \centering
        \includegraphics[width=\linewidth]{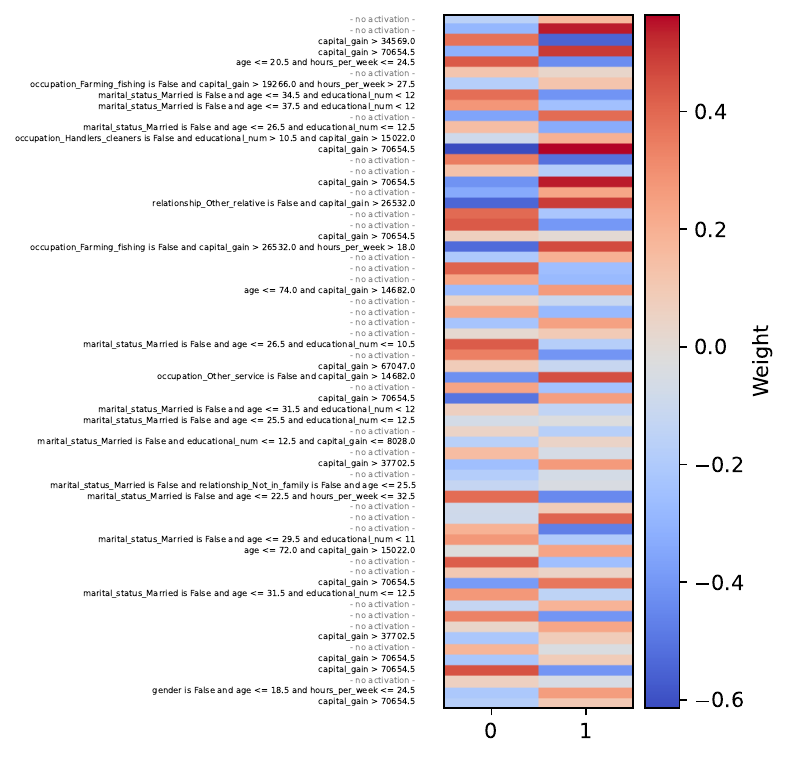}
    \end{minipage}\hfill
    \begin{minipage}[t]{0.5\textwidth}
    \vspace{1em} %
    \hspace{2em} 
    \begin{tabular}{lcc}
    \toprule
    & \dsadult & \dschurn \\\midrule
    \multicolumn{3}{c}{\textsc{Rule Length}}  \\[2pt]
    ~~~min & 1   & 1   \\
    ~~~avg  & 2.18 & 1.8 \\
    ~~~max  & 3   & 2   \\\midrule
    \multicolumn{3}{c}{\textsc{Active Features}} \\[2pt]
    ~~~min & 23   & 19  \\
    ~~~avg & 30   & 33 \\
    ~~~max & 37  & 40  \\\bottomrule
    \end{tabular}  
    \end{minipage}
    \caption{Left: Decision weight matrix $W'$ for \dsadult on interpretable features. Right: Statistics on complexity of features on \dsadult and \dschurn.}
    \label{fig:results:global-expl-adult}
\end{figure}

\textbf{Results.} Table~\ref{tab:results:performance} shows that, on both datasets, \modelname outperforms interpretable models. 
On \dsadult, \modelname is comparable to MLP and slightly below the performance of XGBoost and Rando Forest. On \dschurn, \modelname is on par with all baseline blackbox models including XGBoost. 
These results show that our model introduces interpretability to DNNs with very little compromise on performance. A selection of rules extracted for the dictionary features on \dsadult are shown in Table~\ref{tab:dict_features_adult_selection} (full list in Appendix~\ref{appendix_qualitative_adult}, for \dschurn in Appendix~\ref{appendix_qualitative_churn}). 
The learned rules provide a global explanation of the model's behavior (Figure~\ref{fig:results:global-expl-adult} left), since the predictions are simply a linear combination of the learned features (rules). Each weight in this linear combination directly indicates whether the presence of the corresponding features increases or lowers the probability of a class, and to which extent.
Figure~\ref{fig:results:global-expl-adult} (right) shows that on average 30 out of the 64 features that we learn for \dsadult are active above the chosen threshold.
That means, one local explanation requires us to inspect 30 rules to understand the model's prediction.
The number of active features on average slightly increases for \dschurn, but not the complexity of the rules (average rule length remains 2).
While the number of active rules is high, several rules are redundant and could be pruned. A pruning mechanism could improve the interpretabiliy of the explanations, and while not in the scope of this paper, it is part of our future work.

\begin{table}[ht]
\centering
\caption{Performance of blackbox (\ding{53}) and fully interpretable (\checkmark) models on two benchmark datasets. Best values are marked bold, best values for interpretable models bold italic.}
\label{tab:results:performance}
\begin{tabular}{clcccc}
\toprule
 &   & \multicolumn{2}{c}{\dsadult}& \multicolumn{2}{c}{\dschurn}\\ 
& & F1-Macro& Acc & F1-Macro & Acc\\ \otoprule
\ding{53} & Random Forest 
    &  0.799 $\pm$ 0.003 & 0.861 $\pm$ 0.002 & 0.747 $\pm$ 0.013 & \textbf{ 0.862 $\pm$ 0.007}\\
\ding{53} & XGBoost 
    & \textbf{0.815 $\pm$ 0.002} & \textbf{0.869 $\pm$ 0.001} & 0.750 $\pm$ 0.012 & 0.861 $\pm$ 0.007 \\
\ding{53} & MLP 
    & {0.796 $\pm$ 0.002} & 0.850 $\pm$ 0.002 & 0.751 $\pm$ 0.106 & 0.860 $\pm$ 0.006 \\\midrule
\checkmark & Logistic Regression 
    & 0.782 $\pm$ 0.002 & 0.815 $\pm$ 0.002 & 0.601 $\pm$ 0.006 & 0.808 $\pm$ 0.000\\
\checkmark & Decision Tree 
    & 0.787 $\pm$ 0.001 &  0.851 $\pm$ 0.002 & 0.735 $\pm$ 0.008 & 0.849 $\pm$ 0.006\\\checkmark  & \modelname (ours) 
    & \textit{\textbf{0.795 $\pm$ 0.002}} & \textit{\textbf{0.850  $\pm$ 0.002}} &  \textbf{0.752 $\pm$ 0.008}& \textit{\textbf{0.861 $\pm$ 0.002}} \\
\bottomrule
\end{tabular}
\end{table}

\begin{table}[h]
    \centering
    \footnotesize
    \caption{A selection of dictionary features for the \dsadult dataset. $|T_j|$ - size of the training subset that strongly activate feature $j$. Coverage of the rule reported as number of samples and percentage of samples. Table sorted by $|T_j|$.}
    \label{tab:dict_features_adult_selection}
    \begin{tabularx}{\textwidth}{l l X l}
    \toprule
     j & |T\_j| & Description & Coverage \\
    \midrule
    40 & 15294 & \texttt{marital\_status\_Married is False and educational\_num < 13 and capital\_gain <= 8028.0} & 11625/0.76 \\
     9 & 11968 & \texttt{marital\_status\_Married is False and age <= 37.5 and educational\_num < 12} & 7271/0.61 \\
     8 & 11634 & \texttt{marital\_status\_Married is False and age <= 34.5 and educational\_num < 12} & 6517/0.56 \\
    54 & 10889 & \texttt{marital\_status\_Married is False and age <= 31.5 and educational\_num < 13} & 5953/0.55 \\
    44 &  7457 & \texttt{marital\_status\_Married is False and relationship\_Not\_in\_family is False and age <= 25.5} & 3330/0.45 \\
    45 &  3975 & \texttt{marital\_status\_Married is False and age <= 22.5 and hours\_per\_week <= 32.5} & 1516/0.38 \\
     5 &  1758 & \texttt{age <= 20.5 and hours\_per\_week <= 24.5} & 754/0.43 \\
    34 &  1658 & \texttt{occupation\_Other\_service is False and capital\_gain > 14682.0} & 547/0.33 \\
    26 &  1358 & \texttt{marital\_status\_Widowedand is False and capital\_gain > 14682.0} & 546/0.40 \\
    50 &   982 & \texttt{age <= 72.0 and capital\_gain > 15022.0} & 532/0.55 \\
    63 &   978 & \texttt{gender is Female and age <= 18.5 and hours\_per\_week <= 24.5} & 236/0.24 \\
     7 &   290 & \texttt{occupation\_Farming\_fishing is False and capital\_gain > 19266.0 and hours\_per\_week > 27.5} & 209/0.73 \\
    18 &   226 & \texttt{relationship\_Other\_relative is False and capital\_gain > 26532.0} & 185/0.82 \\
    22 &   225 & \texttt{occupation\_Farming\_fishing is False and capital\_gain > 26532.0 and hours\_per\_week > 18.0} & 183/0.82 \\
     3 &   212 & \texttt{capital\_gain > 34569.0} & 153/0.73 \\

    \bottomrule
\end{tabularx}
\end{table}

\begin{figure}[b]
    \centering
    \includegraphics[width=0.6\linewidth]{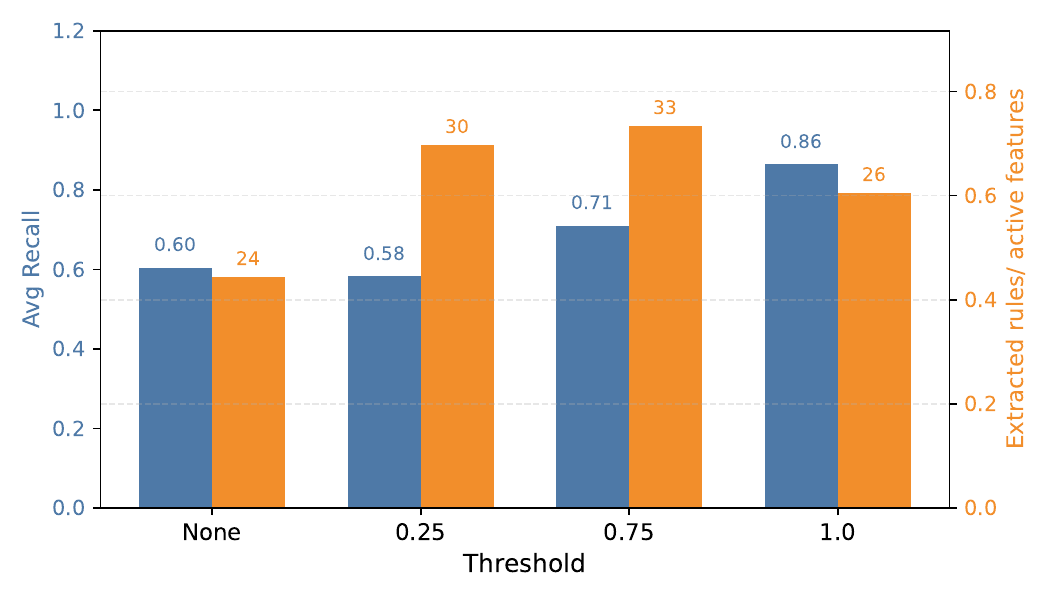}
    \caption{Fraction of features and instances covered by rules to explain \dsadult at different feature activation thresholds $t$.
    \textbf{Orange bars} show the average fraction of rules extracted per active feature with the number of active features on top, \textbf{blue bars} show the average recall of the rules in terms of instances.}
    \label{fig:thresholds}
\end{figure}

\textbf{Impact of activation threshold $t$.}
Figure~\ref{fig:thresholds} shows the impact of $t$ on the number of rules that can be extracted per active feature and on the number of instances covered by each rule (recall). 
The average recall increases with an increasing threshold, as the sets of instances by which features get activated become more focused. This makes it easier for Skope-rules to find a rule that explains most of the instances in the set.
However, a focused set limits the data instances a rule could apply to. For some of these sets, Skope-rules fails to extract meaningful rules, i.e., the sets of instances that are activated by extracted rules become empty. In consequence, the amount of active features for which rules can be extracted decrease with increasing threshold $t$.
Choosing the right threshold requires balancing the trade-off between the average rule recall and the fraction of extracted rules per active features.

\section{Conclusion}
We introduced \modelname, an interpretable model for tabular data. Our results show, that \modelname outperforms interpretable models and performs comparably to blackbox architectures. In future work, we plan to investigate pruning of redundant rules and automated methods to identify activation thresholds for an optimal balance between rule recall and fraction of extracted rules.

\newpage

\clearpage
\section*{Funding Declaration}
This work was funded by the Deutsche Forschungsgemeinschaft (DFG, German Research Foundation) under project ID 536124560.
\bibliographystyle{named}
\bibliography{workshop_ref.bib}
\newpage

\appendix

\section{Implementation Details}
\subsection{Datasets}

\begin{table}[ht]
\centering
\caption{Dataset overview}
\label{tab:appendix:dataset}
\begin{tabular}{ccccc}
\toprule\tabhead
& \# Instances & \# Features & Task \\ \midrule
Adult  & 48 842 &   15  & Binary classification\\
Churn Modeling & 10 000 & 14 & Binary classification\\\bottomrule
\end{tabular}
\end{table}

\textbf{Data Details.} \dsadult  contains demographic information such as age, education, and capital gain. The target variable is wether a person's income is >50k or $\leq 50$.
\dschurn contains details of a bank's customers, the target variable is whether the customer left the bank or continues being a customer.

\textbf{Data preprocessing.} We normalize the input features of \dsadult. For \dschurn, we drop the feature columns ``RowNumber'', ``CustomerId'', and ``Surname'', encode the categorical columns and normalize the numerical ones to $[0,1]$ by min-max normalization. 

\subsection{Parameter settings}
\label{appendix:impl_details:paramter_settings}
For all the models, we use the Optuna library~\footnote{\url{https://optuna.org}} for hyperparameter tuning, the tuning is done on the validation set of each split (5 splits in total). For each experiment, the number of iterations is set to 100. Random seed = 0 

\textbf{MLP.} We use a fixed architecture for both datasets [100, 64, 32]. We tune on the learning rate ($lr$) in the range of [$1e-2$, $5e-3$], dropout rate ($Dropout$) in the range of [0.0, 0.5] and Lambda ($\lambda$) in the range of [$1e-7$, $1e-2$] for L1 regularization.

\textbf{\modelname} Follows the same setting details as MLP for the neural network component, the parameter settings for the sparse autoencoder are the $l1$ coefficient $\alpha = 1e-3$ for \dsadult and \dschurn.

\textbf{Random Forest.} We tuned on number of estimators $(n\_estimators)$ from the list [100, 200, 300], maximum depth $(max\_depth)$ in the range of [5, 10, 15, 20], the minimum samples split $(min\_samples\_split)$ in the range of [2, 10], and the minimum samples leaf $(min\_samples\_leaf)$ in the range of [1, 10].

\textbf{XGBoost.} We set the booster to "gbtree", and "early\_stopping\_rounds" to 12 and tuned on learning rate $(learning\_rate)$ in the range of [0.001, 0.4], maximum depth $(max/_depth)$ in the range of [3, 10], subsample in the range of [0.5, 1], and lambda $(\lambda)$ in the range of [0.1, 10]

\textbf{Decision Trees.} We tuned on maximum depth $(max\_depth)$ in the range of [5, 10, 15, 20], the minimum samples split $(min\_samples\_split)$ in the range of [2, 10], and the minimum samples leaf $(min\_samples\_leaf)$ in the range of [1, 10].

\textbf{Logistic Regression.} We tuned on maximum of iterations $(max\_iter)$ in the range of [100, 200].

For our rule-based classifier:

\textbf{Skope-rules.} we set $precision\_min = 1$, $recall\_min = 0.2$, and the default settings for the rest of the parameters.

\section{Qualitative Results}
\label{appendix_qualitative}
\subsection{\dsadult}
\label{appendix_qualitative_adult}
\begin{table}[h]
    \centering
    \footnotesize
    \caption{Dictionary features for the \dsadult dataset. $|T_j|$ - size of the training subset that strongly activate feature $j$. Coverage of the rule reported as number of samples and percentage of samples. Table sorted by $|T_j|$.}
    \label{tab:dict_features_adult}
    \begin{tabularx}{\textwidth}{l l X l}
    \toprule
     j & |T\_j| & Description & Coverage \\
    \midrule
    40 & 15294 & \texttt{marital\_status\_Married is False and educational\_num < 13 and capital\_gain <= 8028.0} & 11625/0.76 \\
     9 & 11968 & \texttt{marital\_status\_Married is False and age <= 37.5 and educational\_num < 12} & 7271/0.61 \\
     8 & 11634 & \texttt{marital\_status\_Married is False and age <= 34.5 and educational\_num < 12} & 6517/0.56 \\
    54 & 10889 & \texttt{marital\_status\_Married is False and age <= 31.5 and educational\_num < 13} & 5953/0.55 \\
    37 & 10755 & \texttt{marital\_status\_Married is False and age <= 31.5 and educational\_num < 12} & 5735/0.53 \\
    49 &  9304 & \texttt{marital\_status\_Married is False and age <= 29.5 and educational\_num < 11} & 4920/0.52 \\
    11 &  9030 & \texttt{marital\_status\_Married is False and age <= 26.5 and educational\_num < 13} & 4387/0.49 \\
    31 &  8385 & \texttt{marital\_status\_Married is False and age <= 26.5 and educational\_num < 11} & 4098/0.49 \\
    38 &  8313 & \texttt{marital\_status\_Married is False and age <= 25.5 and educational\_num < 13} & 4053/0.48 \\
    44 &  7457 & \texttt{marital\_status\_Married is False and relationship\_Not\_in\_family is False and age <= 25.5} & 3330/0.45 \\
    45 &  3975 & \texttt{marital\_status\_Married is False and age <= 22.5 and hours\_per\_week <= 32.5} & 1516/0.38 \\
     5 &  1758 & \texttt{age <= 20.5 and hours\_per\_week <= 24.5} & 754/0.43 \\
    34 &  1658 & \texttt{occupation\_Other\_service is False and capital\_gain > 14682.0} & 547/0.33 \\
    26 &  1358 & \texttt{marital\_status\_Widowedand is False and capital\_gain > 14682.0} & 546/0.40 \\
    50 &   982 & \texttt{age <= 72.0 and capital\_gain > 15022.0} & 532/0.55 \\
    63 &   978 & \texttt{gender is Female and age <= 18.5 and hours\_per\_week <= 24.5} & 236/0.24 \\
    12 &   521 & \texttt{occupation\_Handlers\_cleaners is False and educational\_num > 10 and capital\_gain > 15022.0} & 404/0.79 \\
    61 &   480 & \texttt{capital\_gain > 70654.5} & 152/0.32 \\
     7 &   290 & \texttt{occupation\_Farming\_fishing is False and capital\_gain > 19266.0 and hours\_per\_week > 27.5} & 209/0.73 \\
    18 &   226 & \texttt{relationship\_Other\_relative is False and capital\_gain > 26532.0} & 185/0.82 \\
    22 &   225 & \texttt{occupation\_Farming\_fishing is False and capital\_gain > 26532.0 and hours\_per\_week > 18.0} & 183/0.82 \\
     3 &   212 & \texttt{capital\_gain > 34569.0} & 153/0.73 \\
    58 &   154 & \texttt{capital\_gain > 37702.5} & 153/0.99 \\
    42 &   153 & \texttt{capital\_gain > 37702.5} & 153/1.00 \\
    36 &   152 & \texttt{capital\_gain > 70654.5} & 152/1.00 \\
     4 &   152 & \texttt{capital\_gain > 70654.5} & 152/1.00 \\
    33 &   152 & \texttt{capital\_gain > 67047.0} & 152/1.00 \\
    53 &   152 & \texttt{capital\_gain > 70654.5} & 152/1.00 \\
    21 &   152 & \texttt{capital\_gain > 70654.5} & 152/1.00 \\
    60 &   152 & \texttt{capital\_gain > 70654.5} & 152/1.00 \\
    16 &   152 & \texttt{capital\_gain > 70654.5} & 152/1.00 \\
    13 &   152 & \texttt{capital\_gain > 70654.5} & 152/1.00 \\
    64 &   152 & \texttt{capital\_gain > 70654.5} & 152/1.00 \\
    \bottomrule
\end{tabularx}
\end{table}

\clearpage
\subsection{\dschurn}
\label{appendix_qualitative_churn}

\begin{table}[ht]
\caption{Dictionary features for the \dschurn dataset. $|T_j|$ - size of the training subset that strongly activate feature $j$. Coverage of the rule reported as number of samples and percentage of samples. Table sorted by $|T_j|$.}
\label{tab:dict_features_churn}
\small
\begin{tabular}{+rr^l^r}
\toprule \tabhead
$j$ & $|T_j|$  & Description & Coverage \\\otoprule
6 & 4955 & \texttt{Age < 45 and NumOfProducts < 3}
    &  4560 / 0.91\\ %
47 & 4631 & \texttt{Age $\leq$ 55.5 and NumOfProducts $\geq$ 2} 
    &  2898 / 0.62\\ %
2 & 4628 & \texttt{CreditScore > 379.0 and Age < 40}
    &  3561 / 0.76\\ %
3 & 4457 & \texttt{CreditScore > 409.0 and NumOfProducts $\geq$ 2} 
    &  2939 /0.66\\ %
30 & 4419 & \texttt{Age < 45 and Balance > 76177.5} 
    &  2594 /0.57\\ %
43 & 3966 & \texttt{Age $\leq$ 42} 
    &  2238 / 0.62\\ %
9 & 2999 & \texttt{Age < 46.5 and NumOfProducts $\geq$ 2}
    &  2533 / 0.84\\ %
12 & 2574 & \texttt{Age < 35 and NumOfProducts $\geq$ 2}
    &  1193 / 0.46\\ %
18 & 1805 & \texttt{Geography\_Spain == False and Age $\geq$ 45}
    &  1081 / 0.59\\ %
36 & 425 & \texttt{Age > 59.5 and  IsActiveMember == 1} 
    &  255 / 0.60\\ %
7 & 424 & \texttt{NumOfProducts $\geq$ 3} 
    &  196 / 0.46\\ %
24 & 363 & \texttt{CreditScore > 361.5 and NumOfProducts $\geq$ 3} 
    &  211 / 0.57\\ %

1 & 350 & \texttt{CreditScore > 379.0 and NumOfProducts $\geq$ 3} 
    &  205 / 0.59\\ %

38 & 303 & \texttt{Age > 64.0 and IsActiveMember == 1} 
    &  172 / 0.57\\ %
28 & 270 & \texttt{Balance $\leq$ 244642.5 and NumOfProducts $\geq$ 3} 
    &  215 /0.79\\ %
40 & 226 & \texttt{NumOfProducts $\geq$ 3}
    &  196 / 0.87\\ %
11 & 218 & \texttt{Age > 62.5 and IsActiveMember == True} 
    &  164 /0.74\\ %
35 & 150 & \texttt{Age $\geq$ 38 and Balance $\geq$ 75720} 
    &  98 /0.65\\ %
18 & 146 & \texttt{Geography\_Germany == True and IsActiveMember == False} 
    &  54 /0.29\\ %
4 & 127 & \texttt{Age > 61.5 and NumOfProducts < 2 and IsActiveMember == True} 
    &  101 / 0.79\\
16 & 112 & \texttt{Age > 66.0 and NumOfProducts < 2 and IsActiveMember == True} 
    &  60 / 0.55\\
5 & 53 & \texttt{Age > 77} 
    &  12 / 0.23\\

\bottomrule
\end{tabular}
\end{table}

\end{document}